\documentclass{article}
\usepackage[utf8]{inputenc}
\usepackage[T1]{fontenc}
\usepackage{graphicx}
\usepackage[top=1in, bottom=1in, left=1in, right=1in]{geometry}
\usepackage{amsmath}
\usepackage{amsfonts}
\usepackage{mathrsfs}

\usepackage{makecell}
\usepackage{color}
\usepackage[toc,page]{appendix}
\usepackage[hidelinks]{hyperref}
\usepackage[english]{babel}
    
\begin{document}

\newcounter{SetCounter}
\setcounter{SetCounter}{\value{footnote}}
\addtocounter{SetCounter}{1}
\setcounter{footnote}{\value{SetCounter}}



\title{Trading USDCHF filtered by Gold dynamics via HMM coupling}
\author{Donny Lee\thanks{Duke University(but not affiliated), Rhicon Currency Management, email: \href{dlee@rhicon.com}{dlee@rhicon.com}} } 
\date{10 June, 2013}

\maketitle

\begin{abstract}
We devise a USDCHF trading strategy using the dynamics of gold as a filter. Our strategy involves modeling both USDCHF and gold using a coupled hidden Markov model (CHMM). The observations will be indicators, RSI and CCI, which will be used as triggers for our trading signals. Upon decoding the modeling in each iteration, we can get the next most probable state and the next most probable observation. Hopefully in taking advantage of intermarket analysis and the Markov property implicit in the model, trading with these most probable values will produce  profitable results. \let\thefootnote\relax


\end{abstract}


\section{Introduction}
Hidden Markov Models (HMM) have their merits in modeling systems. For the past several decades, they found their applications in the area of speech recognition. The underlying principle in HMMs is that we suppose a system is in a certain state at any time. As time progress, the system can transit to other states at probabilities that obey the $markov$ property, which is transitions to another state at time $t+1$ only depends on the state at time $t$ . Its history of states before $t$ is irrelevant.

The reason why the markov model is hidden is because we can only make guesses as to what state the system is in by looking at its output or observations. From there, we use techniques known as $estimation$ or $inference$ to get a best guess of the parameters of the model. Knowning the parameters can then give us the most probable state to which the model will switch enabling us to make predictions of future observations. 

Following the theory of HMMs, research has been made in the formulation of the situation where two different HMMs have correlations in their state progressions. We call this model a coupled Hidden Markov Model (CHMM). Now, the switch of one HMM's state will depend on that HMM's own state and the other HMM's state, of which its own progression will also depend on the itself and the other HMM. As we shall show the specifics later, a rigorous derivation is required to characterize this coupled relationship. The main benefit of CHMMs is that now we can model multiple interacting sequences, a situation which is common in real world phenomena. 

What we wish to do is to develop a trading strategy which involves modeling the currency and commodity market using a CHMM. Here, the hidden states represent the asset's market. Based on how one models the CHMM, the observations can either be prices or technical indicators such as SMA, RSI or Stochastics of the asset. HMM have already found their place in the quantitative trading literature. \cite{TradingHMM} applies HMM to foreign exchange rates and \cite{PredictionHMM} uses HMM to analyze financial time series. It's argued that currency markets, or to a larger extent any financial market, obeys the markov property \cite{MarkovInFinance}. A common issue persistant in any trading strategy is the problem of lag where the meaning of an indicator at present time holds little value as it is based on price action much further in the past that is no longer relevant in determing future movements. The markov property in HMMs solves this problem by design since state transitions only depend of the state before and not on previous states. 

We hope to progress further by analyzing two markets using a CHMM allowing us to capitalize on the CHMM's feature of factoring the correlation between the two. In currency trading, there's a notion that some currencies are correlated with some commodities. Supposing we could characterized this correlation with a CHMM, we hope that in trading the currencies, the information from the commody market will enable us to make better trading decisions. At the same time, we maintain the markov property as was present in the HMM.

Our objective is to first formulate a CHMM and use it to model the movement of USDCHF and of gold and then devise a trading strategy which, by leverging on their coupling, is rooted in accurately predicting future states of both markets.

\section{Formulation of a coupled Hidden Markov model}

As there are many good publications on the formulation of a HMM, we shall not write it here. The formulation we will use is taken from the work done by Shi Zhong and Joydeep Ghosh\cite{mainarticle}. With permission from the authors, I'm reproducing important parts their formulation here with some added features of my own.

\subsection{The CHMM architecture}
There exist several CHMM architectures. The one we will be using is known as the standard fully-coupled HMMs which refers to a group of HMM models in which the state of one model at time $t$ depends on the states of all models (including itself) at time $t+1$. While this architecture can be extended to accommodate any number of HMMs coupled together, we shall henceforth limit our CHMM to containing 2 coupled HMMs. The number of nodes $N$ will be the same for both.

\begin{figure}[h!]
\centering
\includegraphics[scale=0.3]{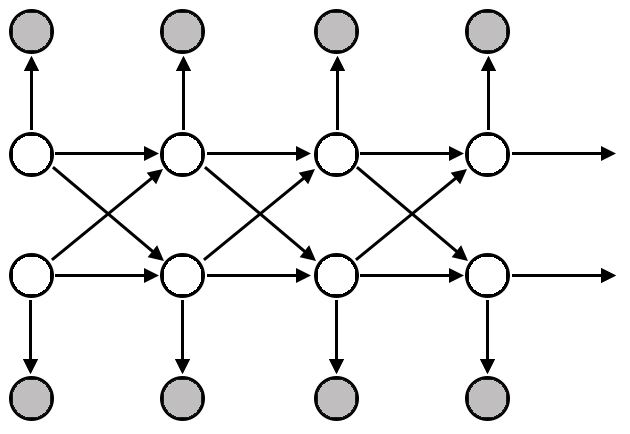}
\caption{Architecture of our coupled hidden markov model}
\label{fig:ChmmFigure}
\end{figure}

Transition probabilities of states for either model now depend on the current state of its HMM and the coupled HMM. So instead of $P(S_t|S_{t-1})$ as in HMMs, we have $P(S_t^{(c)}|S_{t-1}^{(1)},S_{t-1}^{(2)})$. As seen in (Fig ~\ref{fig:ChmmFigure}.), each row of white nodes represent the sequence of each HMM and the grey nodes represent the separate observations emitted from each model in its sequence. 

We'll have 4 transition matrices labeled as $a_{ij}^{(1,1)}$, $a_{ij}^{(1,2)}$, $a_{ij}^{(2,1)}$, $a_{ij}^{(2,2)}$ where $a_{ij}^{(p,q)}$ gives us the probability contribution of transiting to the $j$th node in model $q$ while in the $i$th in $p$th model. Clearly, we have the constraint $\sum_j^N a_{ij}^{(p,q)}=1$ for all $i$, $p$ and $q$.
The joint transition probability to go to a state in either HMM $c$ is

\begin{equation}
P(S_t^{(c)}|S_{t-1}^{(1)},S_{t-1}^{(2)})=\theta_{(1,c)}a_{S_{t-1}^{(1)},S_{t-1}^{(c)}}^{(1,c)} + \theta_{(2,c)}a_{S_{t-1}^{(2)},S_{t-1}^{(c)}}^{(2,c)} 
\end{equation}

where $\theta_{c',c}$ is the coupling weight from HMM $c'$ to HMM $c$, i.e. how much $S_{(t-1)}^{(c')}$ affects $S_{(t-1)}^{(c)}$ by way of the probability $P(S_t^{(c)}|S_{t-1}^{(c')})$. The observation probabilities and prior probabilities are defined as per the standard HMM formulation, but this time we'll have two sets, one for each model, i.e. $b_j^{(1)}(o_t^{(1)})$, $b_j^{(2)}(o_t^{(2)})$, $\pi_j^{(1)}$ and $\pi_j^{(2)}$.

\subsection{Parameter space in the CHMM}

Coupling 2 HMMs gives us the following parameter space.
\newline\newline
The prior probabilities are $\Pi=\left\{ \pi_{j}^{(c)}  \right\}$ where $c=1,2$ and $1\leq j \leq N$ with $\sum_j^N \pi_j^{(c)}=1$.
\newline\newline
The transition probabilities are $A=\left\{ a_{ij}^{(c',c)}  \right\} $ where $c,c'=1,2$ and $1\leq i,j \leq N$ with $\sum_{j}^N a_{ij}^{(c'c)}=1$.
\newline\newline
The observation probabilities are $B=\left\{b_j^{(c)}(k)\right\}$, $c=1,2$ and $1\leq j \leq N$ with $\text{LB}\leq k \leq \text{UB} $ and $\sum_{k=1}^M b_j^{(c)}(k) = 1$.
\newline\newline
The coupling coefficients are $\Theta=\left\{ \theta_{c',c} \right\}$, $c',c=1,2$ with $\theta_{(1,c)}+\theta_{(2,c)} = 1$ for $c=1,2$.
\newline\newline
We will be using a discrete probability distribution\footnote{This is the most primitive distribution. Certainly a harder to parameterized continuous distribution can be used but we'll leave that for future work.} where the outputs are modelled as a range of values. LB and UB defines the lower bound and upper bounds of the output. We will then partition the range into $M$ equal gaps. $b_j^{(c)}(k)$ assigns a probability for each of the gaps. So, for example, if $\text{LB}=0$, $\text{UB}=100$ and $M=4$, $b_j^{(c)}(k)$ will assign probabilities to the outcome intervals $\left\{ (0,25), (25,50), (50,75) , (75,100) \right\}$\footnote{If necessary, we will go into the analysis of whether intervals are closed or opened}.

\subsection{Finding the Viterbi path}

Given a sequence of observations, we might also want to find the most probable hidden path or optimal path that would have output that sequence. We define the optimal path $Q$ as the path that maximizes $P(Q|O,\lambda)$ which is equivalent to maximizing $P(Q,O|\lambda)$. For HMM, there is the $Viterbi\;Algorithm$ to find such a $Q$. We extend its implementation to solve for CHMM.
Instead of considering just one optimal path for a HMM, we need to consider two optimal paths for our 2 model CHMM given two observation sequences. So, we wish to find the optimal paths $Q^{(1)}=\left\{q_1^{(1)},q_2^{(1)},\ldots,q_T^{(1)}\right\}$ and $Q^{(2}=\left\{q_1^{(2)},q_2^{(2)},\ldots,q_T^{(2)}\right\}$ for given observation sequences $O^{(1)}=\left\{o_1^{(1)},o_2^{(1)},\ldots,o_T^{(1)}\right\}$ and $O^{(2)}=\left\{o_1^{(2)},o_2^{(2)},\ldots,o_T^{(2)}\right\}$. Define the quantity

\begin{equation}
\delta_t^{(c)}(i) = \max_{q_1^{(c)},q_2^{(c)},\ldots,q_{t-1}^{(c)}}P\bigg[  q_1^{(c)},q_2^{(c)},\ldots,q_t^{(c)}=s_i^{(c)},o_1^{(c)},o_2^{(c)},\ldots,o_t^{(c)}  |\lambda  \bigg]
\end{equation}

\noindent which means the highest probability along HMM $c$ at time $t$ that output the first $t$ observations and ends in state $s_i^{(c)}$. By inducation we have,

\begin{equation}
\delta_{t+1}^{(c)}(k)=\max_{\substack{ 1 \leq i \leq N \\ 1 \leq j \leq N }}\bigg[\delta_t(i)^{(c)} a_{ik}^{(1,c)} a_{jk}^{(2,c)} \bigg]b_k^{(c)}(o_{t+1}^{(c)})
\end{equation}

To be able to retrieve the optimal paths, we need to keep track of the argument which maximized (3) for each $t$, $(i,j)$ pairing and $c$. We create the array $\psi_t^{(c)}(k)$ to do this. The complete procedure for finding the optimal path for HMM $c$ is as follows:
\newline\newline
1. Initialization:

\begin{equation}
\delta_1^{(c)}(i) = \pi_i^{(c)}b_i^{(c)}(o_1^{(c)}), \;\;\; 1\leq i\leq N
\end{equation}
\begin{equation}
\psi_1^{(c)} = 0 \;\;\;\;\;\;\;\;\;\;\;\;\;\;\;\;\;\;\;\;\;\;\;\;\;\;\;\;\;\;\;\;\;\;\;\;\;\;\;\;\;
\end{equation}
\newline
2. Recursion:

\begin{equation}
\delta_{t}^{(c)}(k)=\max_{\substack{ 1 \leq i \leq N \\ 1 \leq j \leq N }}\bigg[\delta_{t-1}(i)^{(c)} a_{ik}^{(1,c)} a_{jk}^{(2,c)} \bigg]b_k^{(c)}(o_{t}^{(c)}) , \;\;\;\;\;\; 2\leq t \leq T, 1 \leq k \leq N
\end{equation}
\begin{equation}
\psi_{t}^{(c)}(k)=\underset{\substack{ 1 \leq i \leq N \\ 1 \leq j \leq N }}{\operatorname{argmax}}  \bigg[\delta_{t-1}(i)^{(c)} a_{ik}^{(1,c)} a_{jk}^{(2,c)} \bigg], \;\;\;\;\;\;\;\;\;\;\;\;\;\;\;\;\; 2\leq t \leq T, 1 \leq k \leq N
\end{equation}
\newline
3. Termination:

\begin{equation}
P^{(c)}=\max_{1\leq i \leq N}\big[ \delta_T^{(c)}(i)\big]
\end{equation}
\begin{equation}
q_T^{(c)} =  \underset{1\leq i \leq N} { \operatorname{argmax} } \big[ \delta_T^{(c)}(i) \big]
\end{equation}
\newline
4. Optimal path backtracking:
\begin{equation}
q_t^{(c)} = \psi_{t+1}^{(c)}(q_{t+1}^{(c)}), \;\;\;\;\;\;\;\;\;t=T-1,T-2,\ldots,1
\end{equation}

To get the optimal paths for our CHMM, we simply perform the above for HMM 1 and 2.

\subsection{Learning the parameters for CHMM}
As with HMM models, we need to devise a way to learn the parameters of our CHMM from the observations. While we could port the EM/GEM algorithm for HMMs over to CHMMs, doing so will make calculations computationally intractable, at least with the algorithm in its present developed form. Therefore, we'll revert to classical optimization techniques.

The Baum-Welch method used to trained HMMs contains within it theorems which allows us to derive reestimation formulae for the parameters in the CHMM. \cite{mainarticle} has done this nicely to give us the following reesitmation for the model parameters for HMM C:

\begin{equation}
a_{ij}^{(c',c)}=\frac{a_{ij}^{(c',c)}\partial P / \partial a_{ij}^{(c',c)}}{\sum_k^Na_{ik}^{(c',c)}\partial P / \partial a_{ik}^{(c'c)}}
\end{equation}

\begin{equation}
b_{i}^{(c)}(k)=\frac{b_{i}^{(c)}(k)\partial P / \partial b_{i}^{(c)}(k)}{\sum_k^Nb_{i}^{(c)}(k)\partial P / \partial b_{i}^{(c)}(k)}
\end{equation}

\begin{equation}
\pi_{i}^{(c)}=\frac{\pi_{i}^{(c)}\partial P / \partial \pi_{i}^{(c)}}{\sum_k^N\pi_{k}^{(c)}\partial P / \partial \pi_{k}^{(c)}}
\end{equation}

\begin{equation}
\theta_{c'c}=\frac{  \theta_{c'c} \partial P / \partial  \theta_{cc'} }{\sum_{k}^2 \theta_{kc} \partial P / \partial \theta_{kc}}
\end{equation}

Refer to the appendix for the calculation of partial derivatives. It may be unclear how does one HMM affect the reesitmation of parameters for the other HMM since all formulae for reestimating $a_{ij}^{(c',c)}$, $b_i^{(c)}(k)$ and $\pi_i^{(c)}$ for HMM $c$ doesn't take into account the parameters of the other HMM. The dependance on the other HMM's parameters is present in the calculation of $\partial P / \partial a_{ik}^{(c',c)}$, $\partial P / \partial b_{i}^{(c)}(k)$ and $\partial P / \partial \pi_{i}^{(c)}$ as seen in the appendix.

As with learning for HMMs, for each iteration, we use the present HMMs' parameters to estimate new ones for all $A$, $B$, $\Pi$ and $\Theta$ and then swap the present parameters with the new ones.

\section{Strategy setup}
\begin{figure}[h!]
\centering
\includegraphics[scale=0.6]{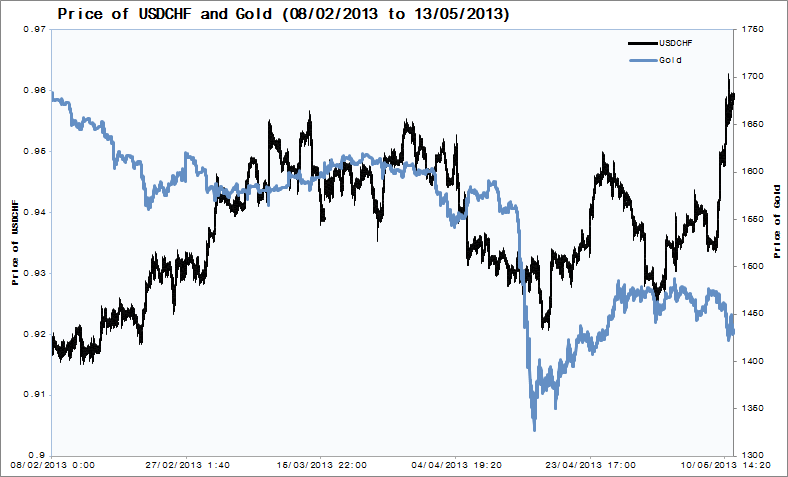}
\caption{Price of USDCHF and gold}
\label{fig:USDCHFGold}
\end{figure}

In order for us to capitalize on the predictive powers of a CHMM, it is best that we use it to model two separate assets that are either strongly correlated or strongly uncorrelated. Facts of currency and commodity trading will help us choose these two assets. One, it is believed that during times of economic unrest, investors tend to dump the dollar in favor of gold. That gold maintains its intrinsic value implies a negative correlation between the US dollar and gold. Two, following a gold selling program, the Swiss National Bank held 1,290 tonnes of gold in reserves which equated to 20\% of Switzerland's assets\cite{swissgold}. Therefore, the Swiss dollar and gold should move in opposite directions. (Fig ~\ref{fig:USDCHFGold}) Both of these facts would have us believe that USDCHF and gold are negatively correlated, the rise in one implies a fall in the other. 

\subsection{ Defining the observation sequence }

USDCHF and gold will be our two assets modelled by the CHMM. The word asset used henceforth can either mean a currency asset or a commodity asset. For both assets, we record down their closing prices. Now, we must be careful in how we define the observation sequence. Initially, one may decide to use the closing prices as the observations. However, we may run into a few problems. First, we will have a hard time finding the range of the observations since by considering prices, we do not have a definite bound on it. From Jan 2013 to Apr 2013, USDCHF has a lower bound of 0.9021 and an upper bound of 0.9568. For gold, it's 1323 and 1702\footnote{Calculated using maximum and minimum periods of the highs and lows in 10 min intervals for the four months}. Due to our limitations of a discrete probability distribution, we have to fix a range for our observations. But we cannot say a prior whether the prices of either will break either of their respective bounds past Apr 2013. Second and more importantly, in order for the transition probabilities $a_{ij}^{(1,2)}$ and $a_{ij}^{(2,1)}$ to have meaning, the degree to which a single unit change of the observation of an asset in one HMM affects the transition probabilities of a specific HMM model must be the same degree as affected by a single unit change of the observation of an asset in the other HMM. These two problems can be easily solved if we normalize the observations to take values in a certain range.

For reasons which will be clearer later, we decide to have our observation sequences be the Relative Strength Index (RSI) of the closing price of USDCHF and of gold. This way, we know that observations from either model will be between 0\% and 100\% and because of that, a percentage change in observations from one HMM will have the same effect as a percentage change in observations from the other HMM. Our choice of basing our indicators on 10-min charts will be explained more later.

Therefore, we define our observation sequences as follows. The RSI of USDCHF as $O^{(1)}=o_1^{(1)},o_2^{(1)},o_3^{(1)},\ldots,o_t^{(1)}$ and the RSI of gold as $O^{(2)}=o_1^{(2)},o_2^{(2)},o_3^{(2)},\ldots,o_t^{(2)}$ where $  0\leq o_t^{(1)},o_t^{(2)}\leq 100  $. We'll use $(1)$ to represent USDCHF's HMM and $(2)$ gold's HMM.

\subsection{Sketch of trading algorithm}

The main idea in deciding when to enter a trade is as follows. 
\newline\newline
1. At the close of a bar, record down the RSI of the price of the last $T$ bars for both assets giving us the observation sequence $O^{(1)}$ and $O^{(2)}$.
\newline
2. With the observation sequence $O^{(1)}$,$O^{(2)}$ find the optimal set of parameters $\lambda=(\pi,A,B,\Theta)$ for our CHMM. This may require us iterate the training process more than once until the increment in $P(O,\lambda)$ falls within a certain threshold.
\newline
3. Based on how we choose to derive the next most probable state, that is $S_{t+1}^{(1)}$ and $S_{t+1}^{(2)}$, find them for both assets.
\newline
4. Using the newly learned parameters $B$ for our observation probabilities, find the most probable output of the most probable observation for both assets.
\newline
5. Define a rule on this observation that will tell us whether to enter a trade.
\newline\newline
Our trading rule would be the simple 4 period Simple Moving Average (SMA) of 4 period RSI overbought and oversold levels. We will go long when SMA of RSI crosses over the level 20\% and short when SMA of RSI crosses under the level 80\%. SMA is used to even out the noise, more so for the RSI values predicted by CHMM since they would be discretized. We'll use a simple take profit level at 6 times the 12 period ATR into the money and a stop loss of 2 times the12 period ATR against the money\footnote{Obviously these rules can be more complicated. Since we are mainly testing the ability of the CHMM to give us good entry signals from its forecasting feature, we are keeping the exits simple.} giving us a 3:1 risk to reward ratio. 
Entries will be at the open using triggers from the previous bar. Exits are at the ATR level.

What follows are four strategies that uses different aspects of the CHMM to generate a trading signal. Since the CHMM is made of several componenets, our goal is to analyze the model and deduce which componenets give a more accurate signal and which ones don't.  
\subsection{Strategy one: next most probable state via transition probabilities}
In finding the most probable state to base our next observation from, we use only the transition matrices to find it. Consider one HMM. We want to find which of the $N$ states each HMM (the destination) is most likely to transit to. By taking transition probability matrics of this own HMM to itself and the other HMM (the source) to this HMM separately, we sum the probabilities of transiting to a particular state on the destination HMM from all the states in the source HMM. Both sums from the two source HMMs are then weighted and added together. The most likeliest state for the target HMM is then the state with the highest sum probabilities. Specifically,

\begin{equation}
\psi^{(1)} = \underset{j}{\arg\max}\left[\theta_{(1,1)}     \sum_i^Na_{i,j}^{(1,1)}       +\theta_{(2,1)}\sum_i^Na_{i,j}^{(2,1)} \right]
\end{equation}
\begin{equation}
\psi^{(2)} = \underset{j}{\arg\max}\left[\theta_{(2,2)}     \sum_i^Na_{i,j}^{(1,1)}       +\theta_{(1,2)}\sum_i^Na_{i,j}^{(1,2)} \right]
\end{equation}
where $\psi^{(1)}$ and $\psi^{(2)}$ is the most likeliest state for USDCHF and gold as calculated in this way. From here, we make predictions using $b_{\psi^{(1)}}^{(1)}$ and $b_{\psi^{(2)}}^{(1)}$.

\subsection{Strategy two: next most probable state via Viterbi path}
Here, $\phi^{(1)}$ and $\phi^{(2)}$ are simply the states at time $T$ of each Viterbi paths. With them, we again use the transition probabilities to predict the next state. Now, the state at time $t$ is fixed but the CHMM will evolve to $\psi^{(1)}$ and $\psi^{(2)}$ where for each HMM, they are determined by the weighted contributions of probability given $\phi^{(1)}$ and $\phi^{(2)}$. Specifically,

\begin{equation}
\psi^{(1)} = \underset{j}{\arg\max}\left[\theta_{(1,1)}     a_{{\phi^{(1)}},j}^{(1,1)}       +\theta_{(2,1)}a_{\phi^{(2)},j}^{(2,1)} \right]
\end{equation}
\begin{equation}
\psi^{(2)} = \underset{j}{\arg\max}\left[\theta_{(2,2)}     a_{{\phi^{(2)}},j}^{(1,1)}       +\theta_{(1,2)}a_{\phi^{(1)},j}^{(1,2)} \right]
\end{equation}

As before, predictions are made using $b_{\psi^{(1)}}^{(1)}$ and $b_{\psi^{(2)}}^{(1)}$.
\subsection{Strategy three: dynamic allocation via transition probabilities}
Now, for each trade, we scale the size of our position based on how probable the next state is as told by the transition probabilities. Regardless of whether we use Viterbi or not to get our next most probable state, we can derive a probability measure from the transition probabilities that tells us the probability of reaching this state as compared to the others. Specifically, given $\psi^{(1)}$, the quantity

\begin{equation}
\frac{ \left[  \sum_{i}^N\sum_{j}^N    \theta_{(1,1)}     a_{i,\psi^{(1)}}^{(1,1)}       +\theta_{(2,1)} a_{j,\psi^{(1)}}^{(2,1)} \right] }{   \left[  \sum_{k}^N\sum_{i}^N\sum_{j}^N    \theta_{(1,1)}     a_{i,k}^{(1,1)}       +\theta_{(2,1)} a_{j,k}^{(2,1)} \right]    }  \;\; \; \;\;\;\;\;\;\;\;\;\;\;\;\;\;(=X)
\end{equation}

is a probability measure and it tells us the probability of switching to state $\psi^{(1)}$ if not for the likeliest state at the current time given by Viterbi. 

We can then use the quantity, a value between 0 and 1, to determine the size of our position. To keep things simple, the scaling is linear. So if our default size per trade is 1,000,000, the new size will be $X$ multiplied by that amount. The reasoning here is that the more likely the HMM will switch to $\psi^{(1)}$, the more capital we want to risk.

\subsection{Strategy four: same as before but using the CCI indicator}
We repeat 3.1 to 3.5 but use a strategy based around the CCI indicator which when backtested produces fairly profitable results. Now, our observation sequence becomes a 4-period CCI. Values for the CCI indicator in 10 mins range from -133 to 133 so the discretization of the observation bound will be eight evenly space intervals from -140 to 140. We long (short) at the open if the previous 4-period SMA of a 4-period CCI crosses under 105 (over -105). Stop loss is a 4 times a 24-period ATR from the trade entry price and take profit is a 10 times a 24-period ATR from the trade entry price. Sine this CCI is used to detect large moves in price action, we will only allow only one long or short position at a time.
\section{Experimental setup}
As with the times when anyone deals with trading strategy, there is the decision of what time interval to use. We are not restricted to a certain time interval, be it 5, 10 or 60 minutes, when recording down our observations for our CHMM. As you shall see later, we ran our strategy using prices in 10 minute intervals.
\subsection{Parameters}

The time frame of data we will be using is OHLC data for USDCHF and gold at 10 minute intervals. Just like how technical indicators can be applied to any time frame \cite{TechIndTime}, our CHMM strategy can also be applied to any time frame. However, we must be careful in selecting the appropriate time frame that gels well with the inherit property of our model, that is the markov property. In selecting a time interval, we are assuming that the state of the market obeys the markov property when it swtiches once every interval. What we are supposing is that all the information required to predict the next state is captured within that interval. The implication here is that too long an interval and we might lost some information which would otherwise be meaningful in determining the next state and too short an interval we deal with just random noise.

Picking an appropriate time frame would be a study in itself. Trading enthusiasts even argue that strategies should involve multiple time frames. For now, we'll stick with our choice of OHLC in 10 mins. 

The choice of parameters for our CHMM are 4 periods for the look back, 5 hidden states and 8 discretization steps given the outcome of observations in steps of $\{0, 12.5, 25, 37.5, 50, 62.5, 75, 87.5, 100\}$. We chose 4 for the look back to tally with our period of 4 for the RSI strategy we will be using. Essentially, we are looking as far back as does the technical indicator\footnote{Choosing more than 4 makes computations more intensive disabling us to get meaningful results in due time.}. The 5 hidden states will represent the states super bear, bear, random, bull and super bull of the economy. Lastly, discretizing the observation range of 0\% to 100\% into 8 steps allows us to nicely specific levels 20\% and 80\% for the signals. A value below 20 for SMA of RSI signifies oversold levels and above 80 overbought levels. In learning the parameters of the CHMM, we repeat updating parameters $\Pi,A,B,\Theta$ three times through each iteration.

\subsection{Data integrity}
We get our 10-min data from the CQG\cite{cqg}. USDCHF and gold's open, high, low, close data were taken from 01/01/2013 to 05/01/2013 specificed by symbols X.US.IUSDCHF and F.US.GCE. Time for calculations depends on the lookback, number of hidden states and discretization parameters of the CHMM. Our current set up was chosen given the computational power and time we worked with.

\section{Backtest results}

\subsection{Performance analysis}

\begin{figure}[h!]
\centering
\includegraphics[scale=0.6]{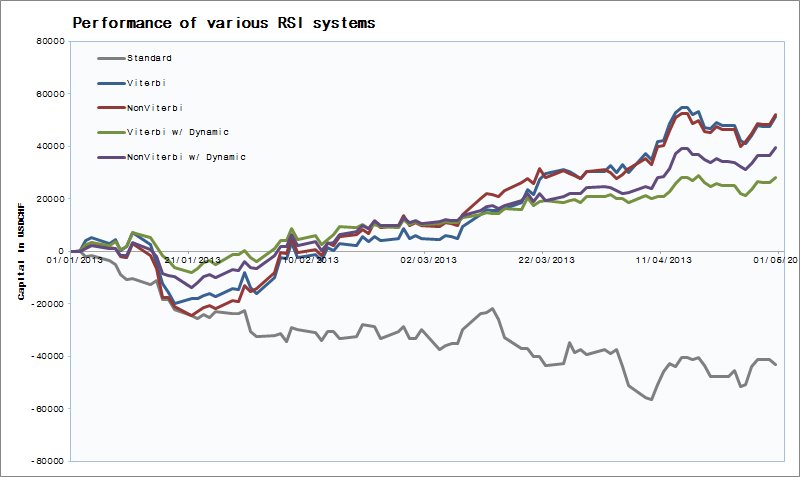}
\caption{Performance of the various RSI systems}
\label{fig:PerfSheet}
\end{figure}

\begin{figure}[h!]
\centering
\includegraphics[scale=0.6]{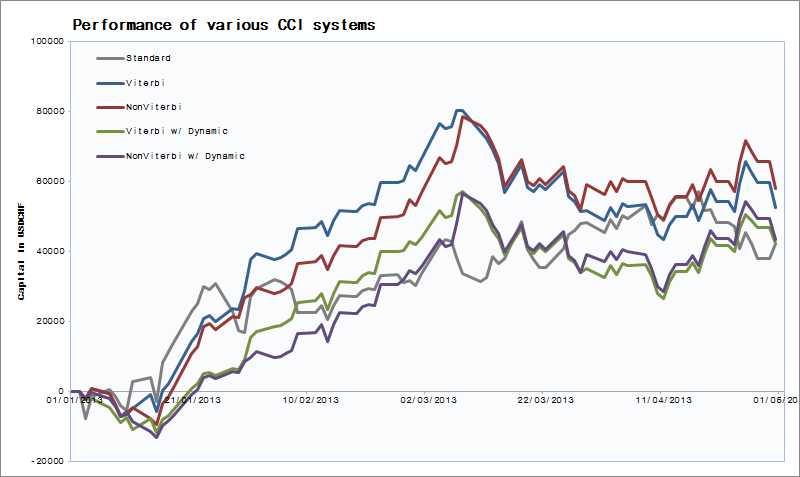}
\caption{Performance of the various CCI systems}
\label{fig:PerfSheetCCI}
\end{figure}

All four systems using CHMM outperformed their standard version for both the RSI and CCI systems. (Fig ~\ref{fig:PerfSheet} and Fig ~\ref{fig:PerfSheetCCI}) Recall that our strategy setup was taking a tested strategy and use the CHMM to predict future values for the indicators used in the system. For RSI, the system was making losses but for CCI, the system was fairly profitable. More importantly, for both systems there was an increase in returns upon using the CHMM.

\begin{figure}[h!]
\centering
\begin{tabular}{l|l|l|l|l}
Name & Ret & Vol & Ratio & $\Delta$ from Ratio \\
\Xhline{2\arrayrulewidth}
RSI Standard & -4.55 & 5.18 & -0.878 & 0 \\
Viterbi & \textbf{5.51} & 6.88 & 0.801 & +1.679 \\
Non Viterbi & 5.09 & 6.18 & 0.824 & +1.702 \\
Viterbi w/ Dynamic & 2.98 & \textbf{2.76} & \textbf{1.080} & \textbf{+1.958} \\
Non Viterbi w/ Dynamic & 3.91 & 3.74 & 1.045 & +1.923 \\
\end{tabular}
\caption{Returns, volatility and it's ratio for the RSI systems. Bold values are best ones for that column.}
\label{fig:PerfTable}
\end{figure}

\begin{figure}[h!]
\centering
\begin{tabular}{l|l|l|l|l}
Name & Ret & Vol & Ratio & $\Delta$ from Ratio \\
\Xhline{2\arrayrulewidth}
CCI Standard & 0.35 & 0.90 & 0.389 & 0 \\
Viterbi & 0.45 & 1.47 & 0.306 & -0.083 \\
Non Viterbi & \textbf{0.49} & 1.15 & 0.426 & +0.037 \\
Viterbi w/ Dynamic & 0.36 & 1.03 & 0.350 & -0.039 \\
Non Viterbi w/ Dynamic & 0.37 & \textbf{0.81} & \textbf{0.457} & \textbf{+0.068} \\
\end{tabular}
\caption{Returns, volatility and it's ratio for the CCI systems. Bold values are best ones for that column.}
\label{fig:PerfTable}
\end{figure}

Between CHMM systems, we see an improvement in the Sharpe ratio\footnote{Risk free returns taken to be 0.0\%} when we use dynamic allocation. Volatility drops for both systems using dynamic allocation while returns also drop. (See Fig ~\ref{fig:PerfTable}) We feel that this is expected because the probability of the next most probable state is largely concentrated in the 40\% to 60\% region (See Fig.~\ref{fig:NextProb}). Given that wihtout dynamic allocation, we always allocate 100\% of our capital, returns are bound to drop.

Our conclusion is that information contained in the CHMM in the form probabilities of the next state does have value in optimizing the trading system. In this case, it helps stabilizes the trading system by decreasing volatility, which is usually a good thing as believed by traders.

\subsection{Viterbi analysis}

Now, we do a quick analysis on how much different the next probable state is when using Viterbi and non-Viterbi methods. As the performance for both differs, but only  slightly, we expect that the probable states was predicted by both methods will also differ slightly.

One practical use in knowing that the Viterbi and non-Viterbi methods are quite faithful to each other is that we could abandone the more computational intensive one in favor for the less computational intensive one.

\section{Conclusion}
Based on the performance of the trading system of RSI overbought oversold where one system uses the actual RSI value and four others use the RSI value as predicted by the CHMM, we see that the latter drastically outperforms the former. The reasons are twofold. One, our CHMM introduced additional data, the dynamics of gold, into our trading strategy. More importantly, CHMM allows a convenient coupling, via the transition matrix, that positively affects trading performance using this new information. Two, implicit in the model is the markov property which is believed to remove the problem of lag pronounced in the RSI strategy.

Of course, right now, it is only a theory that CHMMs give better indicators for generating trade signals. Further substantiating this theory requires us to look at the myriad of trading systems common in trading literature, i.e. Stochastics, MACD divergence, MA crossovers, employ a CHMM with an appropriate cross market asset and with its observations being the indicators for the system conconcerned, and then compare performance figures. We hope that by doing so, performance figures will improve based on the beneficial features of the CHMM.

\section{Future developments}
\begin{figure}[h!]
\centering
\includegraphics[scale=0.38]{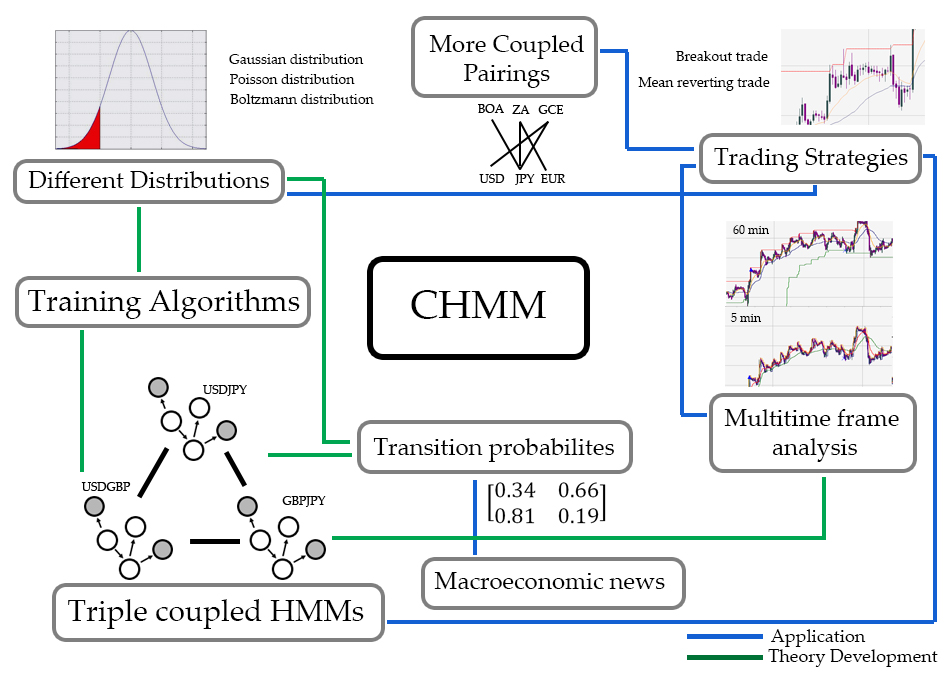}
\caption{Workplan for our CHMM trading engine}
\label{fig:WorkPlan}
\end{figure}

The results of this paper has made us to believe the value in CHMM used in formulating a profitable trading strategy. We like its markov feature and its ability to link HMMs of different markets giving us a more confident trading signal. Due to its merits, we hope to develop an entire trading engine centered around CHMM as seen in Fig ~\ref{fig:WorkPlan}.

\subsection{ Other trading strategies involving CHMM }

For this paper, we used a CHMM to predict the next RSI value of the assets and conclude whether the asset is overbought or oversold. In similar fashion, we can use  CHMM to forecast values of other indicators and use them to generate trading signals. For example, we might want to forecast Stochastic oscillator and use it to filter out false breakout levels. The use of CHMM acts as a complement to improve established breakout systems. 

Or for example, we can use the forecasted indicators of the both assets and determine a signal based on how far they diverge or converge in the style of a mean reverting strategy. The supposed correlation of the two assets is accounted for by the inherit coupling between the states. 

\subsection{ Different representation of the observation functions }

Our work only dealt with a discrete probability distribution where the range of values is discretized into regularly spaced intervals. Having the observations take a continuous distribution introduces a wealth of new problems we need to tackle - what distribution to use, the learning of parameters, weights for mixtures.

One hopeful benefit in working with continuous distributions is the improved accuracy in forecasting as this distribution better represents the distribution of observations. However, the added cost in using continuous distributions is the computational workload we incur during the learning of parameters of the distribution. And for more complication distributions, how we learn the paramters may not be a trivial task in itself.

\subsection{ Three coupled HMMs  }
A nifty idea, yet one very far in the workplan, is to explore three coupled HMMs modeling triangular currency pairs, i.e. USDJPY, GBPJPY and USDGBP. In currency trading, there's  opportunities in such triple linked pairs. One such opportunity is the Triangular Arbitrage. While we aren't considering arbitrage style strategies, we can still leverage on the coupling feature of CHMMs to see whether there are good strategies involving such three currencies.

\subsection{Multitime frame analysis}
CHMM, if done right, can also be used for multitime frame trading strategies. We could use the coupling of HMMs to investigate a relationship between the macro picture of the charts, say 60 min OHLC, and the micro picture of the charts, say two regions of 5 min OHLC. In formulating a trading strategy, a breakout as seen in 5 min could be linked to how price action evolved in 60 min or vice versa. 
\newline\newline
We believe that by carefully studying each area of the CHMM starting with the ones listed above, we will reach closer to the goal of developing a very profitable trading system. 



\newpage
\appendix
\noindent\chapter{\LARGE{\textbf{Appendix}}}
\numberwithin{equation}{section}
\section{Chart of CHMM indicators}
What follows are the charts of various values outputed by the CHMM. To keep charts uncluttered, we show portions of the time series in Feb 2013.

\begin{figure}[h!]
\centering
\includegraphics[scale=0.85]{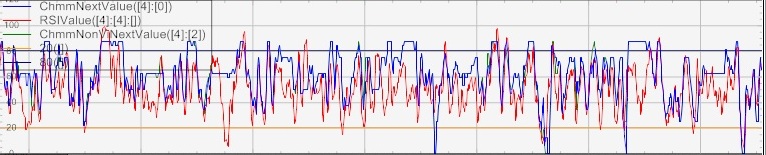}
\caption{Most probable RSI value from the next most probable state using Viterbi and Non-Viterbi methods.}
\label{fig:NextValue}
\end{figure}

\begin{figure}[h!]
\centering
\includegraphics[scale=0.85]{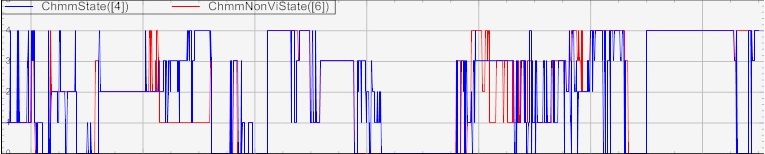}
\caption{Next most probable state using Viterbi and Non-Viterbi methods.}
\label{fig:NextState}
\end{figure}

\begin{figure}[h!]
\centering
\includegraphics[scale=0.85]{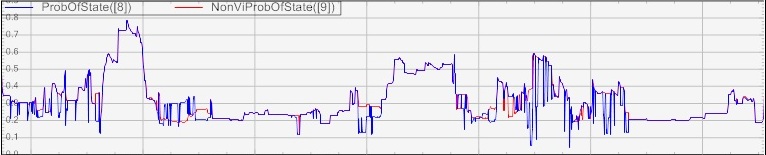}
\caption{The probability of going to the next most probable state using Viterbi and Non-Viterbi methods.}
\label{fig:NextProb}
\end{figure}

\section{Partial derivatives}
\cite{mainarticle} details the calculation patrial Derivatives for reestimation of CHMM. We merely relist them here. Let $\vec{w}=(\pi,A,B,\Theta)$ be the parameter vector. Define

\begin{equation}
\delta_{x,y} =
\begin{cases} 
1, & \text{if } x=y \\ 
0, & \text{if } x \neq y 
\end{cases}
\end{equation}
\begin{equation}
z_{ijt}^{(c',c)}=\theta_{c',c}\cdot a_{ij}^{(c',c)} \cdot b_j^{(c)}(o_t^{(c)})
\end{equation}

We have

\begin{equation}
\alpha_t^{(c)}(j)=
\begin{cases}
\pi_j^{(c)} \cdot b_j^{(c)}(o_1^{(c)}), & t=1 \\
\sum_{c'}\sum_i z_{ijt}^{(c',c)}\alpha_{t-1}^{(c')}(i), & 2 \leq t \leq T
\end{cases}
\end{equation}

\begin{equation}
\frac{\partial P}{ \partial w } = \sum_c \left( \frac{P}{P^{(c)}} \frac{\partial P^{(c)}}{ \partial w } \right) = \sum_c \left( \frac{P}{P^{(c)}} \sum_{j=1}^N \frac{ \partial \alpha_T^{(c)} (j) }{ \partial w } \right)
\end{equation}

The first order derivatives of $\alpha_t^{(c)}(j)$ with respect to each type of parameters are as follows.

For $\partial \alpha_t^{(c)}(j) / \partial \pi_i^{(c_1)}$,

\begin{equation}
\frac{\partial\alpha_t^{(c)}(j)}{\partial\pi_t^{(c_1)}} =
\begin{cases}
\delta_{ij}\delta_{c,c_1}\cdot b_j^{(c_1)}(o_1^{(c_1)}), & t=1 \\
\sum_{c'=1}^{C}\sum_{k=1}^{N(c')}z_{kjt}^{(c',c)}\frac{\partial\alpha_{t-1}^{(c')}}{\partial\pi_i^{(c_1)}}, & 2 \leq t \leq T
\end{cases}
\end{equation}

For $\partial\alpha_t^{(c)}(j)/\partial a_{ij}^{(c',c)}$,

\begin{equation}
\frac{\partial\alpha_t^{(c)}(j)}{\partial a_{ij}^{(c_1,c_2)}} =
\begin{cases}
0, & t=1 \\
\delta_{c,c_2}\delta_{j,j_1}\theta_{c_1c_2}b_{j_1}^{(c_2)}(o_t^{(c_2)}\alpha_{t-1}^{(c_2)}(i_1))+\sum_{c'}\sum_i z_{ijt}^{(c',c)}\frac{\partial\alpha_{t-1}^{(c')}(i)}{\partial\alpha_{i_1 j_1}^{(c_1,c_2)}}, & 2 \leq t \leq T
\end{cases}
\end{equation}

For $\partial\alpha_t^{(c)}(j)/\partial b_{j_1}^{(c_1)}(k)$,

\begin{equation}
\frac{\partial\alpha_t^{(c)}(j)}{\partial b_{j_1}^{(c_1)}(k)} =
\begin{cases}
\delta_{o_1^{(c)},k} \delta_{c,c_1} \delta_{j,j_1} \pi _{j_1}^{c_1}                                 , & t=1 \\
\sum_{c'}\sum_i\left( \delta_{ o_1^{(c)},k }\delta_{c,c_1}\delta_{j,j_1} \theta_{c'c_1} a _{ij_1}^{(c,c_1)} \alpha_{t-1}^{(c')}(i) + z_{ijt}^{(c',c)} \frac{\partial\alpha_{t-1}^{(c')}(i)}{\partial b_{j_1}^{(c_1)}(k)}   \right)                          , & 2 \leq t \leq T
\end{cases}
\end{equation}

For $\partial\alpha_t^{(c)}(j)/\partial \theta_{c_1c_2}$,

\begin{equation}
\frac{\partial\alpha_t^{(c)}(j)}{\partial\theta_{c_1c_2}} =
\begin{cases}
0,                                       & t=1 \\
\delta_{c,c_2}\sum_i a_{ij}^{(c_1,c_2)}b_j^{(c_2)}(k)\alpha_{t-1}^{(c_1)}(i)+\sum_{c'}\sum_i z_{ijt}^{(c',c)}\frac{\partial\alpha_{t-1}^{(c')}(i)}{\partial \theta_{c_1,c_2}},                       & 2 \leq t \leq T
\end{cases}
\end{equation}

\end{document}